\documentclass[11pt]{article}
\usepackage[margin=1in]{geometry}
\usepackage{amsmath,amssymb}
\usepackage{graphicx}
\usepackage{booktabs}
\usepackage{hyperref}
\usepackage{natbib}
\usepackage{xcolor}
\usepackage{tikz}
\usepackage{pgfplots}
\usepackage{multirow}
\pgfplotsset{compat=1.18}
\usetikzlibrary{arrows.meta,positioning,calc,fit,backgrounds}

\title{Position-Agnostic Pre-Projection for Transformer Attention:\\Nonlinear Feature Construction and Content Skip Before Q/K/V}

\author{
Chirag Shinde\\
\textit{Independent Researcher}\\
\texttt{chirag.m.shinde@gmail.com}
}

\date{}

\begin{document}
\maketitle

\begin{abstract}
We propose two complementary modifications to transformer attention blocks. First, a nonlinear \textit{pre-projection} MLP is inserted between layer norm and Q/K/V projections, constructing richer features in a position-agnostic manner before any positional encoding is applied. Second, a \textit{content skip} connection routes the pre-projection's features around the attention mechanism, allowing content information to bypass position-aware attention where beneficial. In frozen-probe experiments on Pythia-160M and 410M, the combined approach achieves the strongest results across methods: \textbf{+40.6\% LAMBADA accuracy} and \textbf{$-$39\% perplexity} at 160M scale. Learned skip connection weights reveal a consistent pattern across model sizes: later transformer layers activate the content bypass more strongly than earlier layers, suggesting that deeper layers benefit from content information that does not pass through positional attention. All modifications add no K/V cache overhead.
\end{abstract}

\section{Introduction}

Modern transformer blocks compute attention through a fixed pipeline: layer normalization, linear Q/K/V projections, rotary positional encoding (RoPE) on Q and K, then scaled dot-product attention. This design has two limitations we address:

\textbf{Linear feature bottleneck.} The Q/K/V projections are purely linear. If useful attention patterns depend on nonlinear feature combinations, the model must rely on previous layers' FFNs to have constructed them---requiring multiple layers for what could be a single-layer operation.

\textbf{Mandatory positional filtering.} All information reaching the output must pass through position-aware attention. Content information that is relevant regardless of position (semantic type, syntactic category, entity identity) is forced through the same positional routing as position-dependent information.

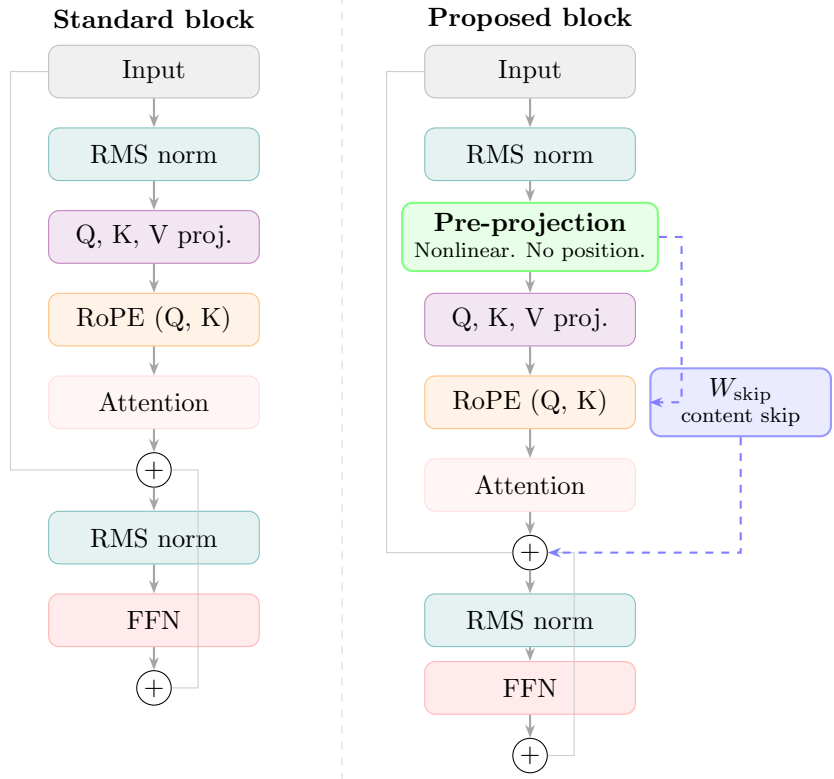
\begin{figure}[t]
\centering
\begin{tikzpicture}[
    box/.style={draw, rounded corners=4pt, minimum width=2.8cm, minimum height=0.7cm, align=center, font=\small},
    graybox/.style={box, fill=gray!12, draw=gray!50},
    tealbox/.style={box, fill=teal!10, draw=teal!50},
    purplebox/.style={box, fill=violet!10, draw=violet!50},
    amberbox/.style={box, fill=orange!10, draw=orange!50},
    pinkbox/.style={box, fill=pink!15, draw=pink!50},
    coralbox/.style={box, fill=red!8, draw=red!30},
    greenbox/.style={box, fill=green!10, draw=green!50, minimum width=3.4cm, line width=0.8pt},
    bluebox/.style={box, fill=blue!8, draw=blue!40, minimum width=2.4cm, line width=0.8pt},
    arr/.style={-{Stealth[length=5pt]}, gray!70, thick},
    resarr/.style={gray!40, thin},
    skiparr/.style={-{Stealth[length=5pt]}, blue!50, thick, dashed},
    >=Stealth
]

\node[font=\small\bfseries] at (-2.5, 8.2) {Standard block};
\node[graybox] (l-in) at (-2.5, 7.5) {Input};
\node[tealbox] (l-norm1) at (-2.5, 6.4) {RMS norm};
\node[purplebox] (l-qkv) at (-2.5, 5.3) {Q, K, V proj.};
\node[amberbox] (l-rope) at (-2.5, 4.2) {RoPE (Q, K)};
\node[pinkbox] (l-attn) at (-2.5, 3.1) {Attention};
\node[draw, circle, inner sep=1pt, font=\small] (l-add1) at (-2.5, 2.2) {$+$};
\node[tealbox] (l-norm2) at (-2.5, 1.3) {RMS norm};
\node[coralbox] (l-ffn) at (-2.5, 0.2) {FFN};
\node[draw, circle, inner sep=1pt, font=\small] (l-add2) at (-2.5, -0.7) {$+$};
\foreach \a/\b in {l-in/l-norm1, l-norm1/l-qkv, l-qkv/l-rope, l-rope/l-attn, l-attn/l-add1, l-add1/l-norm2, l-norm2/l-ffn, l-ffn/l-add2}
    \draw[arr] (\a) -- (\b);
\draw[resarr] (l-in.west) -- ++(-0.5,0) |- (l-add1.west);
\draw[resarr] (l-add1.east) -- ++(0.35,0) |- (l-add2.east);

\node[font=\small\bfseries] at (2.5, 8.2) {Proposed block};
\node[graybox] (r-in) at (2.5, 7.5) {Input};
\node[tealbox] (r-norm1) at (2.5, 6.4) {RMS norm};
\node[greenbox] (r-pre) at (2.5, 5.3) {\textbf{Pre-projection}\\[-2pt]{\scriptsize Nonlinear. No position.}};
\node[purplebox] (r-qkv) at (2.5, 4.2) {Q, K, V proj.};
\node[amberbox] (r-rope) at (2.5, 3.1) {RoPE (Q, K)};
\node[pinkbox] (r-attn) at (2.5, 2.0) {Attention};
\node[draw, circle, inner sep=1pt, font=\small] (r-add1) at (2.5, 1.1) {$+$};
\node[tealbox] (r-norm2) at (2.5, 0.2) {RMS norm};
\node[coralbox] (r-ffn) at (2.5, -0.7) {FFN};
\node[draw, circle, inner sep=1pt, font=\small] (r-add2) at (2.5, -1.6) {$+$};
\foreach \a/\b in {r-in/r-norm1, r-norm1/r-pre, r-pre/r-qkv, r-qkv/r-rope, r-rope/r-attn, r-attn/r-add1, r-add1/r-norm2, r-norm2/r-ffn, r-ffn/r-add2}
    \draw[arr] (\a) -- (\b);
\draw[resarr] (r-in.west) -- ++(-0.5,0) |- (r-add1.west);
\draw[resarr] (r-add1.east) -- ++(0.35,0) |- (r-add2.east);

\node[bluebox] (r-skip) at (5.3, 3.1) {$W_{\text{skip}}$\\[-2pt]{\scriptsize content skip}};
\draw[skiparr] (r-pre.east) -- ++(0.3,0) |- (r-skip.west);
\draw[skiparr] (r-skip.south) |- (r-add1.east);

\draw[dashed, gray!30] (0, 8.5) -- (0, -2.0);
\end{tikzpicture}
\caption{Standard block (left) vs.\ proposed block (right). The green pre-projection constructs nonlinear features before Q/K/V. The blue content skip routes those features around attention, adding directly to the output. Both paths are position-agnostic; positional encoding enters only at RoPE.}
\label{fig:architecture}
\end{figure}

We propose two modifications (Figure~\ref{fig:architecture}):
\begin{enumerate}
    \item A \textbf{pre-projection}: a small nonlinear MLP before Q/K/V that constructs richer, position-agnostic features.
    \item A \textbf{content skip}: a learned linear projection from the pre-projection output that bypasses attention and adds directly to the layer output.
\end{enumerate}

The pre-projection enriches features for attention; the content skip lets those features also bypass attention where the model learns it is beneficial. Both are position-agnostic and add no K/V cache overhead.

\section{Method}

\subsection{Pre-Projection}

Between layer normalization and Q/K/V projections, we insert:
\begin{align}
    \hat{x} &= \text{RMSNorm}(x) \\
    \tilde{x} &= \hat{x} + W_{\text{down}} \cdot \sigma(W_{\text{up}} \cdot \hat{x})
\end{align}
where $W_{\text{up}} \in \mathbb{R}^{d \times ed}$, $W_{\text{down}} \in \mathbb{R}^{ed \times d}$, $\sigma$ is SiLU, and $e = 1.25$. The residual connection ensures near-identity initialization.

\subsection{Content Skip}

The same enriched features $\tilde{x}$ are projected through a learned linear map $W_{\text{skip}} \in \mathbb{R}^{d \times d}$ and added to the output after attention:
\begin{align}
    Q, K, V &= \tilde{x}W_Q,\; \tilde{x}W_K,\; \tilde{x}W_V \\
    \text{attn\_out} &= \text{Attention}(\text{RoPE}(Q), \text{RoPE}(K), V) \\
    \text{out} &= x + \text{attn\_out} + W_{\text{skip}} \cdot \tilde{x}
\end{align}

$W_{\text{skip}}$ is initialized near-zero ($\mathcal{N}(0, 10^{-4})$) so the model starts identical to baseline. The skip learns per-layer how much content information should bypass positional attention. No separate mixing parameter (alpha) is needed---the skip projection's weight magnitude serves this role naturally and trains via standard backpropagation.

\textbf{Key insight}: The pre-projection does double duty. It enriches Q/K/V inputs (position-aware path) \textit{and} provides content features that bypass attention (position-agnostic path). The model learns to route content information through whichever path is more useful at each layer.

\subsection{Parameter Overhead}

\begin{table}[h]
\centering
\caption{Parameter breakdown for the full method (pre-projection + content skip, $e = 1.25$).}
\label{tab:overhead}
\begin{tabular}{lccccc}
\toprule
\textbf{Model} & \textbf{Base} & \textbf{Pre-proj} & \textbf{Skip} & \textbf{Total new} & \textbf{Overhead} \\
\midrule
Pythia-160M & 162M & 17.7M & 7.1M & 24.8M & 13.2\% \\
Pythia-410M & 405M & 62.9M & 25.2M & 88.1M & 17.9\% \\
\bottomrule
\end{tabular}
\end{table}

Neither the pre-projection nor the content skip adds K/V cache overhead. Both are fixed-cost per-token transformations that occur before and after the attention computation respectively.

\section{Experimental Setup}

\subsection{Models, Data, and Benchmarks}

We evaluate on Pythia-160M (12 layers) and Pythia-410M (24 layers) \citep{biderman2023pythia}, training on WikiText-103 \citep{merity2016pointer}. Benchmarks: LAMBADA \citep{paperno2016lambada} (discourse comprehension), HellaSwag \citep{zellers2019hellaswag} (commonsense reasoning), ARC-Easy \citep{clark2018arc} (science QA), and WikiText-103 perplexity.

\subsection{Protocol}

All pretrained parameters are frozen. Only injected parameters train (pre-projection, skip, or LoRA). AdamW, cosine schedule, LR $10^{-5}$, effective batch 16, bf16, 500 steps. Evaluations on 500-sample subsets. LoRA \citep{hu2021lora} baselines use rank chosen to approximately match parameter counts.

\section{Results}

\subsection{Main Results}

\begin{table}[h]
\centering
\caption{Frozen probe results. Base model frozen throughout. \textbf{Bold} = best per model per metric. Trainable parameter counts vary due to rank granularity (LoRA) and skip projection ($d^2$ per layer).}
\label{tab:main_results}
\begin{tabular}{llccccc}
\toprule
\textbf{Model} & \textbf{Method} & \textbf{Params} & \textbf{LAMBADA}$\uparrow$ & \textbf{HSwag}$\uparrow$ & \textbf{ARC-E}$\uparrow$ & \textbf{PPL}$\downarrow$ \\
\midrule
\multirow{5}{*}{\shortstack[l]{Pythia\\160M}}
 & Baseline          & ---   & 0.128 & 0.326 & 0.358 & 78.4 \\
 & Pre-proj          & 17.7M & 0.154 & 0.318 & 0.342 & 73.0 \\
 & LoRA ($r$=480)    & 26.5M & 0.126 & 0.326 & 0.346 & 63.8 \\
 & Pre-proj + LoRA   & 21.2M & 0.162 & 0.324 & 0.352 & 60.8 \\
 & \textbf{Pre-proj + skip}  & 24.8M & \textbf{0.180} & \textbf{0.338} & 0.348 & \textbf{47.9} \\
\midrule
\multirow{5}{*}{\shortstack[l]{Pythia\\410M}}
 & Baseline          & ---   & 0.466 & 0.396 & 0.436 & 29.2 \\
 & Pre-proj          & 62.9M & 0.470 & 0.392 & \textbf{0.452} & 25.1 \\
 & LoRA ($r$=640)    & 94.4M & \textbf{0.488} & 0.400 & 0.446 & 17.7 \\
 & Pre-proj + LoRA   & 72.4M & 0.470 & \textbf{0.400} & 0.450 & 19.2 \\
 & \textbf{Pre-proj + skip}  & 88.1M & 0.484 & 0.398 & 0.432 & \textbf{17.0} \\
\bottomrule
\end{tabular}
\end{table}

Table~\ref{tab:main_results} presents the full comparison. The pre-projection with content skip achieves the strongest overall results:

\textbf{At 160M scale}, pre-proj + skip achieves the best LAMBADA (0.180, \textbf{+40.6\%} over baseline), the best HellaSwag (0.338, +3.7\%), and the best perplexity (47.9, \textbf{$-$38.9\%}). This is the only method that improves HellaSwag---all other methods leave it flat or slightly degrade it.

\textbf{At 410M scale}, pre-proj + skip achieves the best perplexity (17.0), surpassing even standalone LoRA with 94.4M parameters (17.7). LAMBADA (0.484) exceeds all methods except standalone LoRA (0.488, which uses more parameters).

\subsection{Complementarity: Pre-Projection, LoRA, and Content Skip}

The three techniques address different limitations:
\begin{itemize}
    \item \textbf{Pre-projection}: nonlinear feature construction before attention.
    \item \textbf{LoRA}: linear projection enrichment within attention.
    \item \textbf{Content skip}: content bypass around attention.
\end{itemize}

At 160M scale, LoRA alone \textit{decreases} LAMBADA below baseline (0.126) while improving perplexity (63.8)---it optimizes surface prediction without improving comprehension. The pre-projection alone improves comprehension (LAMBADA 0.154) but modestly improves perplexity. The content skip combines both strengths: it achieves the best comprehension \textit{and} perplexity simultaneously.

\subsection{Layer-Wise Content Skip Analysis}

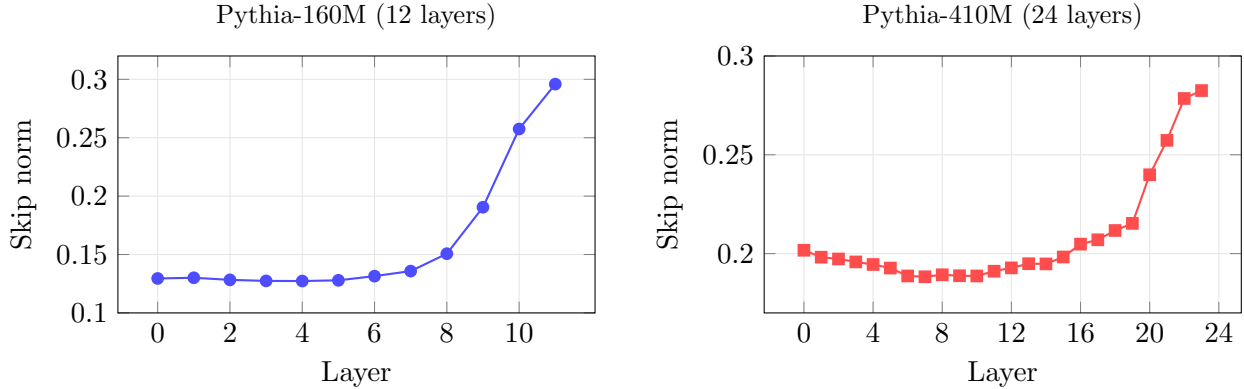
\begin{figure}[t]
\centering
\begin{tikzpicture}
\begin{axis}[
    width=0.48\textwidth, height=5cm,
    xlabel={Layer}, ylabel={Skip norm},
    title={\small Pythia-160M (12 layers)},
    ymin=0.1, ymax=0.32,
    xtick={0,2,4,6,8,10},
    grid=major, grid style={gray!20},
    mark size=2pt,
]
\addplot[color=blue!70, thick, mark=*] coordinates {
    (0,0.1295) (1,0.1301) (2,0.1283) (3,0.1274) (4,0.1273) (5,0.1279)
    (6,0.1315) (7,0.1358) (8,0.1507) (9,0.1905) (10,0.2575) (11,0.2959)
};
\end{axis}
\end{tikzpicture}
\hfill
\begin{tikzpicture}
\begin{axis}[
    width=0.48\textwidth, height=5cm,
    xlabel={Layer}, ylabel={Skip norm},
    title={\small Pythia-410M (24 layers)},
    ymin=0.17, ymax=0.30,
    xtick={0,4,8,12,16,20,24},
    grid=major, grid style={gray!20},
    mark size=2pt,
]
\addplot[color=red!70, thick, mark=square*] coordinates {
    (0,0.2017) (1,0.1982) (2,0.1973) (3,0.1958) (4,0.1945) (5,0.1927)
    (6,0.1887) (7,0.1883) (8,0.1893) (9,0.1888) (10,0.1887) (11,0.1911)
    (12,0.1928) (13,0.1949) (14,0.1948) (15,0.1983) (16,0.2048) (17,0.2070)
    (18,0.2117) (19,0.2153) (20,0.2399) (21,0.2573) (22,0.2785) (23,0.2825)
};
\end{axis}
\end{tikzpicture}
\caption{Learned content skip projection weight norms by layer. Both models show the same pattern: later layers activate the content bypass more strongly. Skip weights are initialized at $\mathcal{N}(0, 10^{-4})$; the growth is entirely learned.}
\label{fig:skip_norms}
\end{figure}

Figure~\ref{fig:skip_norms} reveals a consistent pattern across both model sizes: \textbf{the content skip is activated most strongly in the final layers}. At 160M, skip norm grows from 0.13 (layers 0--7) to 0.30 (layer 11)---a 2.3$\times$ increase. At 410M, skip norm grows from 0.19 (layers 0--10) to 0.28 (layer 23)---a 1.5$\times$ increase. The growth is concentrated in the final 25\% of layers at both scales.

This pattern has a natural interpretation. Early transformer layers are building representations through progressive refinement---positional attention is doing useful routing work, moving information between tokens based on their positions. Later layers have already formed rich semantic representations and are preparing for output prediction. At this stage, content information that bypasses positional attention is valuable: the model has already decided \textit{what} matters, and additional positional routing would only dilute the signal.

This finding connects to prior observations about transformer layer roles \citep{tenney2019bert}: early layers handle syntax (position-dependent), while later layers handle semantics (content-dependent). The content skip learns this division automatically---it activates where content matters more than position.

\section{Discussion}

\subsection{Three Dimensions of Attention Enrichment}

Our experiments reveal three orthogonal ways to improve attention in a frozen probe setting, each addressing a different limitation:

\begin{enumerate}
    \item \textbf{Nonlinear feature construction} (pre-projection): surfaces features invisible to linear Q/K/V projections. Improves comprehension (LAMBADA) most.
    \item \textbf{Linear projection enrichment} (LoRA): makes Q/K/V projections higher-rank. Improves perplexity most.
    \item \textbf{Content bypass} (skip): lets content information avoid positional attention where unhelpful. Improves both comprehension and perplexity.
\end{enumerate}

The content skip combines the benefits of (1) and (2) because the pre-projection's enriched features serve double duty: they improve what goes into attention \textit{and} provide content that bypasses it.

\subsection{Position-Agnostic Design for Multimodal Architectures}

Neither the pre-projection nor the content skip contains positional encoding. Positional information enters only at RoPE on Q/K. This separation of \textbf{content preparation} (modality-independent) from \textbf{positional awareness} (modality-specific) is particularly relevant for multimodal models where text is sequential, images are 2D spatial, video is spatiotemporal, and point clouds have no natural ordering.

In joint-attention architectures like MMDiT \citep{esser2024scaling}, a shared pre-projection and content skip could construct modality-independent features, with each modality applying its own positional encoding only at the Q/K stage. The content skip would be especially valuable for modalities where position is undefined---content features could bypass positional attention entirely.

\subsection{Limitations}

Our experiments use a frozen probe on two model sizes with 500-sample evaluations. Scaling to 1B+ models and from-scratch training remain untested. The LoRA comparison is not perfectly parameter-matched. HellaSwag and ARC-Easy differences are within sampling noise at 500 samples for some methods. The layer-wise skip pattern, while consistent across both tested sizes, should be verified at larger scale.

\section{Related Work}

Talking-heads attention \citep{shazeer2020talking} mixes logits across heads. LoRA \citep{hu2021lora} adds low-rank updates to projections. Adapters \citep{houlsby2019parameter} insert bottleneck layers between sub-layers. Prefix tuning \citep{li2021prefix} prepends trainable tokens. Highway networks \citep{srivastava2015highway} use learned gating for skip connections, though not in the context of bypassing attention. Our work is distinguished by (1) placing nonlinear feature construction specifically before Q/K/V in a position-agnostic manner, (2) providing a learned content bypass around attention, and (3) demonstrating that the bypass activates preferentially in later layers.

\section{Conclusion}

We present a nonlinear pre-projection before Q/K/V projections with a content skip connection that bypasses attention. The pre-projection constructs richer features for attention; the skip lets those features also avoid positional routing where the model learns it is unhelpful. Across two model sizes and four benchmarks, the combined approach achieves the strongest results: +40.6\% LAMBADA accuracy and $-$38.9\% perplexity at 160M scale, with the best perplexity at 410M (17.0, surpassing LoRA's 17.7 with fewer effective LoRA parameters). The learned skip weights reveal a consistent architectural insight: content bypass is most valuable in the final layers, where semantic representations are mature and positional routing becomes less essential. Both modifications add no K/V cache overhead and are position-agnostic by design.

\paragraph{Reproducibility.} Code available at \texttt{https://github.com/cs-cmyk/preprojection}.

\bibliographystyle{plainnat}

\begin{thebibliography}{14}

\bibitem[Biderman et~al., 2023]{biderman2023pythia}
Biderman, S., et~al. Pythia: A suite for analyzing large language models across training and scaling. \emph{ICML}, 2023.

\bibitem[Clark et~al., 2018]{clark2018arc}
Clark, P., et~al. Think you have solved question answering? Try ARC, the AI2 reasoning challenge. \emph{arXiv:1803.05457}, 2018.

\bibitem[Esser et~al., 2024]{esser2024scaling}
Esser, P., et~al. Scaling rectified flow transformers for high-resolution image synthesis. \emph{ICML}, 2024.

\bibitem[Houlsby et~al., 2019]{houlsby2019parameter}
Houlsby, N., et~al. Parameter-efficient transfer learning for NLP. \emph{ICML}, 2019.

\bibitem[Hu et~al., 2022]{hu2021lora}
Hu, E.~J., et~al. LoRA: Low-rank adaptation of large language models. \emph{ICLR}, 2022.

\bibitem[Li and Liang, 2021]{li2021prefix}
Li, X.~L. and Liang, P. Prefix-tuning: Optimizing continuous prompts for generation. \emph{ACL}, 2021.

\bibitem[Merity et~al., 2016]{merity2016pointer}
Merity, S., et~al. Pointer sentinel mixture models. \emph{arXiv:1609.07843}, 2016.

\bibitem[Paperno et~al., 2016]{paperno2016lambada}
Paperno, D., et~al. The LAMBADA dataset: Word prediction requiring a broad discourse context. \emph{ACL}, 2016.

\bibitem[Shazeer et~al., 2020]{shazeer2020talking}
Shazeer, N., et~al. Talking-heads attention. \emph{arXiv:2003.02436}, 2020.

\bibitem[Srivastava et~al., 2015]{srivastava2015highway}
Srivastava, R.~K., Greff, K., and Schmidhuber, J. Highway networks. \emph{arXiv:1505.00387}, 2015.

\bibitem[Tenney et~al., 2019]{tenney2019bert}
Tenney, I., Das, D., and Pavlick, E. BERT rediscovers the classical NLP pipeline. \emph{ACL}, 2019.

\bibitem[Touvron et~al., 2023]{touvron2023llama}
Touvron, H., et~al. LLaMA: Open and efficient foundation language models. \emph{arXiv:2302.13971}, 2023.

\bibitem[Zellers et~al., 2019]{zellers2019hellaswag}
Zellers, R., et~al. HellaSwag: Can a machine really finish your sentence? \emph{ACL}, 2019.

\end{thebibliography}

\end{document}